\documentclass[conference]{IEEEtran}

\usepackage[T1]{fontenc}
\usepackage[utf8]{inputenc}
\usepackage{cite}
\usepackage{amsmath,amssymb}
\usepackage{graphicx}
\usepackage{booktabs}
\usepackage{microtype}
\usepackage[hidelinks]{hyperref}

\title{Embedded Evaluation of Task Admission Coalescing in Decentralized Multi-Robot Systems}

\author{
\IEEEauthorblockN{James Lott\\ \small Student Member, IEEE}
\IEEEauthorblockA{\textit{Shiley Marcos School of Engineering}\\
\textit{University of San Diego}\\
San Diego, United States of America\\
jlott@sandiego.edu}
\and
\IEEEauthorblockN{Vahraz Honary\\ \small Senior Member, IEEE}
\IEEEauthorblockA{\textit{Shiley Marcos School of Engineering}\\
\textit{University of San Diego}\\
San Diego, United States of America\\
vhonary@sandiego.edu}
}

\begin{document}

\maketitle

\begin{abstract}
Multi-robot task allocators in dynamic missions commonly admit newly released tasks immediately, potentially invoking allocation for each new arrival. We evaluate task-admission coalescing as a mechanism for controlling allocator processor work while accounting for target-service latency, using CBAA, ACBBA, PI, and HIPC as a representative MRTA suite. The study comprises a 3,000-mission AGX Orin campaign with measured computation delay, 3,000 paired zero-compute missions, and a 96-mission Pololu 3pi+ RP2040 allocator hardware-in-the-loop campaign. Immediate (Eager) admission is compared with thresholds of two, four, and eight tasks and a four-task policy with a 10-s waiting bound across three arrival rates. Across the AGX experiments, coalescing reduces both allocator calls and processor work in 31 of 48 evaluated conditions, although the magnitude of the saving varies by allocator. The principal cost appears under sparse arrivals, where four-task batching increases online-target mean latency by 20.08--23.57~s, predominantly through admission waiting. Bounded admission reduces mean work in ten of twelve allocator--load conditions, including a 19.35\% reduction with a 1.40-s mean latency increase for high-arrival HIPC. RP2040 experiments show that this tradeoff can become more favorable as processor constraints tighten. Across four matched medium-arrival HIPC scenarios, Count $b=4$ reduces mean RP2040 work by 59.85\% and service latency by 39.47\%, while the corresponding AGX cases increase latency by 27.69\%. These results show that the usefulness of task-admission coalescing depends on arrival intensity, allocator-specific behavior, and the relative cost of computation on the execution platform.
\end{abstract}


\section{Introduction}
\label{sec:introduction}

Decentralized multi-robot task allocation (MRTA) distributes assignment decisions among robots using local state and exchanged peer information~\cite{choi2009consensus,johnson2010asynchronous}. In dynamic missions, new tasks may be released while robots are already executing earlier assignments, requiring the allocation process to incorporate additional work without interrupting ongoing execution. Most online allocation methods react as newly released tasks become known to the team, so successive arrivals can repeatedly introduce new information into an active coordination process.

Prior work has addressed the cost of these repeated updates by reducing the amount of coordination or recomputation required after new information arrives. Johnson et al. extend asynchronous bundle allocation with replanning in response to changes in local situational awareness~\cite{johnson2011dynamic}, while Buckman et al. limit replanning to selected portions of an existing CBBA allocation and can restrict the participating team~\cite{buckman2019partial}. Tuck et al. retain learned information between allocations of newly arriving tasks through incremental satisfiability-modulo-theories solving~\cite{tuck2024smt}. Other approaches reduce coordination frequency rather than replanning scope. Sao et al. suppress CBBA broadcasts when bid and winner information has not changed~\cite{sao2025event}, and Aznar et al. initiate negotiation in response to significant events while adapting the pace of coordination~\cite{aznar2026event}. Together, these approaches show that the computational and communication burden of repeatedly responding to dynamic information is itself an important part of online task allocation.

A related idea appears in systems that deliberately wait before dispatching newly available work. Pavone et al. study robotic routing under stochastic service requests and show that system behavior depends strongly on request arrival load~\cite{pavone2011adaptive}. In online warehouse order batching, Gil-Borr\'{a}s et al. show that the choice of waiting window can materially affect operational performance across batching and routing methods~\cite{gilborras2024window}. These results establish that delaying the release of accumulated work to a downstream decision process can be useful, but they do not address the computational response of a decentralized MRTA system whose robots continue coordinating and executing while new tasks arrive.

This work introduces a task admission layer between external task release and allocator visibility. The layer does not change when tasks arise. Instead, it controls when already released tasks become available to the decentralized allocators. Immediate admission exposes each task as it arrives, while task admission coalescing allows successive releases to be introduced together. This creates a direct tradeoff between the waiting imposed before allocation and the computational activity generated after admission. The outcome is not determined by admission frequency alone because robots continue processing consensus information, repairing allocation state, and selecting work between admissions, while the size and timing of a batch can change both the number and cost of later allocator transactions.

Our preceding Collaborative Visit benchmark provides additional motivation for examining this interaction. That study showed substantial differences in computational burden among decentralized allocators while evaluating a fixed set of known targets under ideal and degraded communication~\cite{lott2026benchmark}. Its execution-integrated timing accumulated allocator work without allowing measured computation to advance mission time. The companion bundle-length study examined how retained planning depth changes MinSum and MinMax behavior under communication loss, again with a fixed task set~\cite{lott2026bundle}. Neither study considered tasks released during execution, waiting before allocator admission, or the feedback that arises when allocator computation consumes mission time. More broadly, Collaborative Visit provides a common setting for comparing decentralized allocation mechanisms under execution and communication effects~\cite{cao2025cv}.

The present study evaluates whether task admission coalescing can reduce the computational burden of dynamic task arrivals without imposing excessive target-service delay. CBAA, ACBBA, PI, and HIPC provide a representative suite spanning different decentralized coordination mechanisms. Immediate, count-based, and bounded admission are evaluated across three arrival rates using paired NVIDIA Jetson AGX Orin experiments with and without causal computation delay and RP2040 allocator hardware-in-the-loop execution.

The study makes three contributions.
\begin{enumerate}
\item A paired characterization of how task admission coalescing changes allocator processor work and target-service latency across four decentralized MRTA mechanisms, five admission policies, and three arrival loads.
\item An analysis of how admission waiting, allocator-call frequency, and per-call work shape the computation--service-latency tradeoff after newly released tasks enter allocation.
\item A measured AGX and RP2040 comparison showing how tighter processor constraints can shift the computation and service-latency tradeoff and, in some operating conditions, allow coalescing to improve both outcomes.
\end{enumerate}

Source code, experimental data, manifests, and analysis materials are available at the accompanying GitHub repository: \url{https://github.com/jlott22/Task_Coalescing_MRTA}.

\section{System Design and Admission Policies}
\label{sec:system}

\subsection{Dynamic Collaborative Visit}
\label{sec:dynamic_cv}
The study extends the fixed-task Collaborative Visit framework in~\cite{lott2026benchmark} to tasks released during execution. Four robots operate on an obstacle-free $19\times19$ grid and must visit 50 distinct task cells. Eight tasks are released initially and 42 arrive online. For task $j$, let $r_j$, $a_j$, and $f_j$ denote its release, environmental admission, and first completion times. Release makes a task available to the admission policy. Admission announces its location to the robots, after which a robot with that local knowledge can complete it by searching the cell. Crossing a task cell before learning its admission does not complete the task.

Each robot maintains its own task knowledge, allocation state, and received peer information. The environment records releases and physical completions but does not select robot assignments. A common robot shell provides four-neighbor A* routing, movement, and communication. The mission ends when every task has been completed. This separates the externally controlled timing of task availability from decentralized decisions about task ownership and visit order.

\subsection{Admission Policies}
\label{sec:admission}
Newly released tasks enter a pending queue ordered by release time and task index. All policies admit the initial eight tasks together at time zero. Eager admits each subsequent task immediately on release. Count admits exactly $b$ pending tasks whenever the queue reaches $b\in\{2,4,8\}$, leaving any smaller remainder pending. Bounded uses $b=4$ and additionally admits the pending remainder when its oldest task has waited $W=10$~s. The deadline limits residence in the environmental queue, rather than subsequent communication, allocation, or service time.

Count requires a terminal rule because 42 online tasks leave a two-task remainder for $b=4$ and $b=8$. This remainder is admitted only after no future release remains and all previously admitted tasks have physically completed. The same rule applies to Bounded when it precedes the timeout. An admission epoch records one admitted batch, including the initial batch. On receipt of an admission announcement, a robot records the new tasks and requests allocation at its next safe control boundary. Its current traversal, valid goal, ownership, and remaining route are preserved.

\subsection{MRTA Allocator Suite}
\label{sec:allocators}
The suite combines four coordination mechanisms under common Manhattan-distance allocation costs. The Consensus-Based Auction Algorithm (CBAA)~\cite{choi2009consensus} maintains one owned task and retains its auction-time winning bid during movement. The Asynchronous Consensus-Based Bundle Algorithm (ACBBA)~\cite{johnson2010asynchronous} constructs bundles through marginal route-insertion bids and timestamped consensus, releasing a lost task and its dependent suffix. Performance Impact (PI)~\cite{zhao2016impact} uses marginal route significance, removes completed or lost items, and reevaluates the remaining path. Hybrid Information and Plan Consensus (HIPC)~\cite{johnson2016hipc} constructs a local team plan from received peer information and reconciles its own path through plan consensus.

The three multi-task implementations consider the full eligible, locally known task set. Admission preserves valid state, while completion and conflict repair follow each implementation's own rules. Allocator calls also arise from delivered consensus information, completion, invalidation, and local recovery, according to allocator specific protocols. A robot without actionable information sleeps without repeated allocator polling. 

\subsection{Execution Architecture}
\label{sec:causal}
\begin{figure}[t]
\centering
% Architecture figure retained from the supplied manuscript.
\includegraphics[width=\columnwidth]{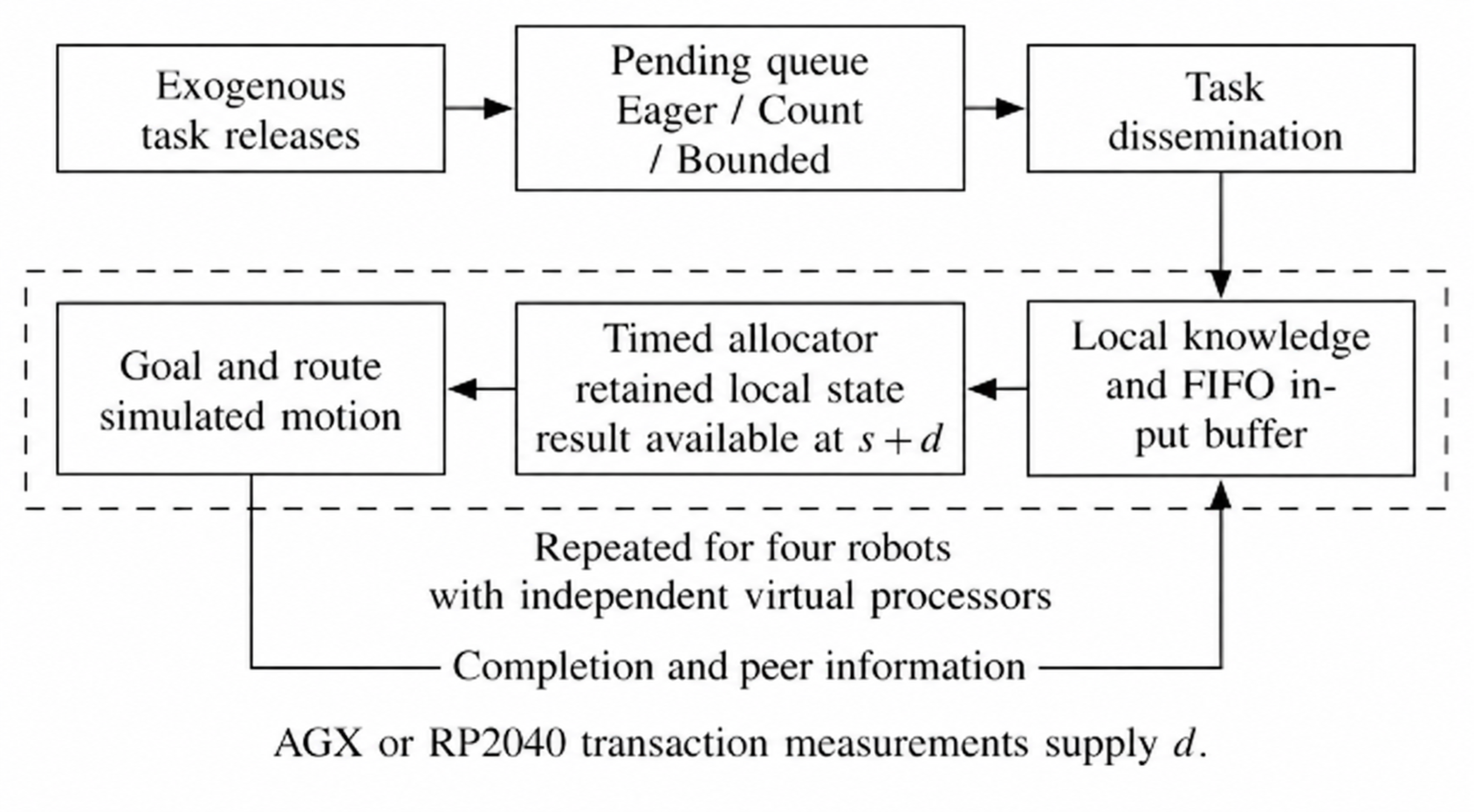}
\caption{Task admission and causal execution architecture. Measured AGX or RP2040 allocator durations advance the corresponding logical robot's virtual processor.}
\label{fig:architecture}
\end{figure}

Each logical robot has an independent virtual processor on the common mission clock. Inputs to simultaneous allocator calls are frozen before execution. A measured transaction duration $d$ delays the calling robot from virtual start $s$ until $s+d$, when its result becomes available. The robot finishes an existing traversal before allocating and does not move while its allocator is busy. Peer activity, releases, and message deliveries continue on the event clock. Information received during computation is buffered for a subsequent transaction. Figure~\ref{fig:architecture} shows the admission and execution boundaries.

The zero-compute arm executes the same host allocator but inserts zero virtual computation time. Its decisions are recomputed from its own evolving state rather than replayed from the causal arm. The comparison therefore captures changes in execution and coordination caused by enabling computation delay.

\section{Experimental Methodology}
\label{sec:methodology}

\subsection{Workloads and Campaign Design}
\label{sec:workloads}
Fifty scenario blocks are generated using separate deterministic spatial, arrival, and runtime streams. Robots start at $(0,0)$, $(0,6)$, $(0,12)$, and $(0,18)$, facing east. Task cells are sampled uniformly without replacement from the 357 non-start cells. For each scenario, 42 independent unit-exponential increments $e_k$ generate online release times
\begin{equation}
 r_{8+k}=\frac{1}{\lambda}\sum_{\ell=1}^{k}e_\ell,
 \qquad k=1,\ldots,42,
 \label{eq:release}
\end{equation}
rounded to nine decimal places. Low, medium, and high arrival loads use $\lambda=0.075$, $0.30$, and $0.60$ tasks/s. The spatial layout and underlying increments are shared across loads, and each scenario is reused across policies, allocators, and timing arms.

Cardinal traversal takes $1.60\pm0.10$~s with uniform jitter, and quarter turns take 0.30~s modeled on testbed measurements in \cite{mypaper1}. Collision-intent settling and replanning take 0.10 and 0.30~s. All messages are delivered at 40~ms in simulations, modeling transmission and processing.  

The AGX design crosses four allocators, three loads, five policies, and 50 scenarios, producing 3,000 missions per timing arm in two 25-scenario rounds. The RP2040 subset uses the first four scenario blocks with Eager and Count $b=4$, producing 96 missions. Table~\ref{tab:campaigns} reports the retained coverage. Runs terminate on completion or at finite limits of 251,000 processed events and 5,500 events without a change in the recorded progress signature. The 60 causal terminations comprise 59 ACBBA stagnation-limit outcomes and one HIPC total-event-limit outcome. 

\begin{table}[t]
\caption{Experimental campaigns and retained coverage.}
\label{tab:campaigns}
\centering
\small
\setlength{\tabcolsep}{3.3pt}
\begin{tabular}{@{}lrrrrr@{}}
\toprule
Campaign & Policies & Traces & Planned & Complete & Other\\
\midrule
AGX causal & 5 & 50 & 3,000 & 2,940 & 60\\
AGX zero-compute & 5 & 50 & 3,000 & 3,000 & 0\\
RP2040 HIL & 2 & 4 & 96 & 82 & 14\\
\bottomrule
\end{tabular}\par
\vspace{3pt}
\noindent\parbox{\columnwidth}{}
\end{table}

\subsection{Platforms and Allocator Measurement}
\label{sec:platforms}
AGX measurements use CPython 3.10.12 on Linux/aarch64 with five concurrent simulation workers. Four Pololu 3pi+ 2040 boards execute native MicroPython allocator transactions. Each board handles one simulated four-robot mission at a time by restoring the relevant robot checkpoint before each call. The host retains world state, routing, motion, and message scheduling, while the boards measure allocator execution without physical robot movement.

The timed boundary includes ordered admission, decoded consensus and completion callbacks, requested local recovery, and goal selection or route repair. AGX uses a monotonic nanosecond timer and RP2040 uses wrap-safe microsecond timing, with naturally triggered garbage collection inside the transaction included. Transport, decoding, checkpoint restoration, explicit pre-call collection, output serialization, logging, routing, and motion are excluded. Physical calls may execute serially for measurement while their durations are assigned to independent virtual robot processors.

\subsection{Outcomes and Paired Analysis}
\label{sec:analysis}
For scenario $t$, let $n_{i,t}$ be robot $i$'s number of valid allocator transactions and $d_{i,t,k}$ the measured duration of transaction $k$. Total allocator processor work and invocation count are
\begin{equation}
 C_{\Sigma,t}=\sum_i\sum_{k=1}^{n_{i,t}}d_{i,t,k},
 \qquad N_{\mathrm{call},t}=\sum_i n_{i,t}.
 \label{eq:work}
\end{equation}
Mean work per call is $\bar c_t=C_{\Sigma,t}/N_{\mathrm{call},t}$, calculated within each mission before averaging. Work counts full dispatched transactions, including those whose virtual completion extends beyond the final task completion. It is accumulated allocator work rather than mission delay or total onboard computation.

Target service latency is the time from release to first completion. Its mean over all 50 targets in a completed scenario is
\begin{equation}
 \bar L_t=\frac{1}{50}\sum_{j=1}^{50}(f_{j,t}-r_{j,t}).
 \label{eq:latency}
\end{equation}
The separately labeled $\bar L_{\mathrm{online},t}$ averages only the 42 online targets. Their release-to-admission waiting $a_j-r_j$ and post-admission time $f_j-a_j$ decompose service latency. Admission count $E_{\mathrm{adm},t}$ supports interpretation of the workload response.

Continuous policy comparisons retain scenarios completed under both the policy and Eager within the same allocator, load, and platform. Planned and completed counts remain visible separately. With $\bar Y_p$ and $\bar Y_E$ denoting means over this common cohort, we report absolute paired differences and percentage changes
\begin{equation}
 Q_p(Y)=100\left(\frac{\bar Y_p}{\bar Y_E}-1\right).
 \label{eq:relative}
\end{equation}
AGX uncertainty uses 10,000 shared scenario-bootstrap resamples~\cite{efron1979bootstrap} and percentile 95\% pointwise intervals. These describe individual contrasts rather than simultaneous coverage across the matrix. Each draw retains the paired outcomes and completion masks across conditions. Tasks, calls, and repeated appearances of a scenario are not separate replicates. Hardware summaries show individual pairs and descriptive changes on matched scenario subsets. The detailed HIPC and CBAA examples were selected from the observed matrix to explain contrasting responses rather than to provide held-out validation.

For the selected mechanism comparisons, the exact within-scenario identity
\begin{equation}
 \Delta C_\Sigma=
 \Delta N\frac{\bar c_p+\bar c_E}{2}
 +\Delta\bar c\frac{N_p+N_E}{2}
 \label{eq:decomposition}
\end{equation}
separates call-count and per-call-work contributions before averaging, with $N=N_{\mathrm{call}}$ and $\Delta$ denoting policy minus Eager. This is an accounting decomposition. Supporting checks use zero-compute call counts and latency over the first 40 online tasks to examine timing feedback and terminal-remainder effects. Diagnostic replay timing is not treated as new AGX or RP2040 measurement.

\section{Results and Discussion}
\label{sec:results}

\begin{figure*}[t]
\centering
% Main policy tradeoff figure retained without changes to plotted data.
\includegraphics[width=\textwidth]{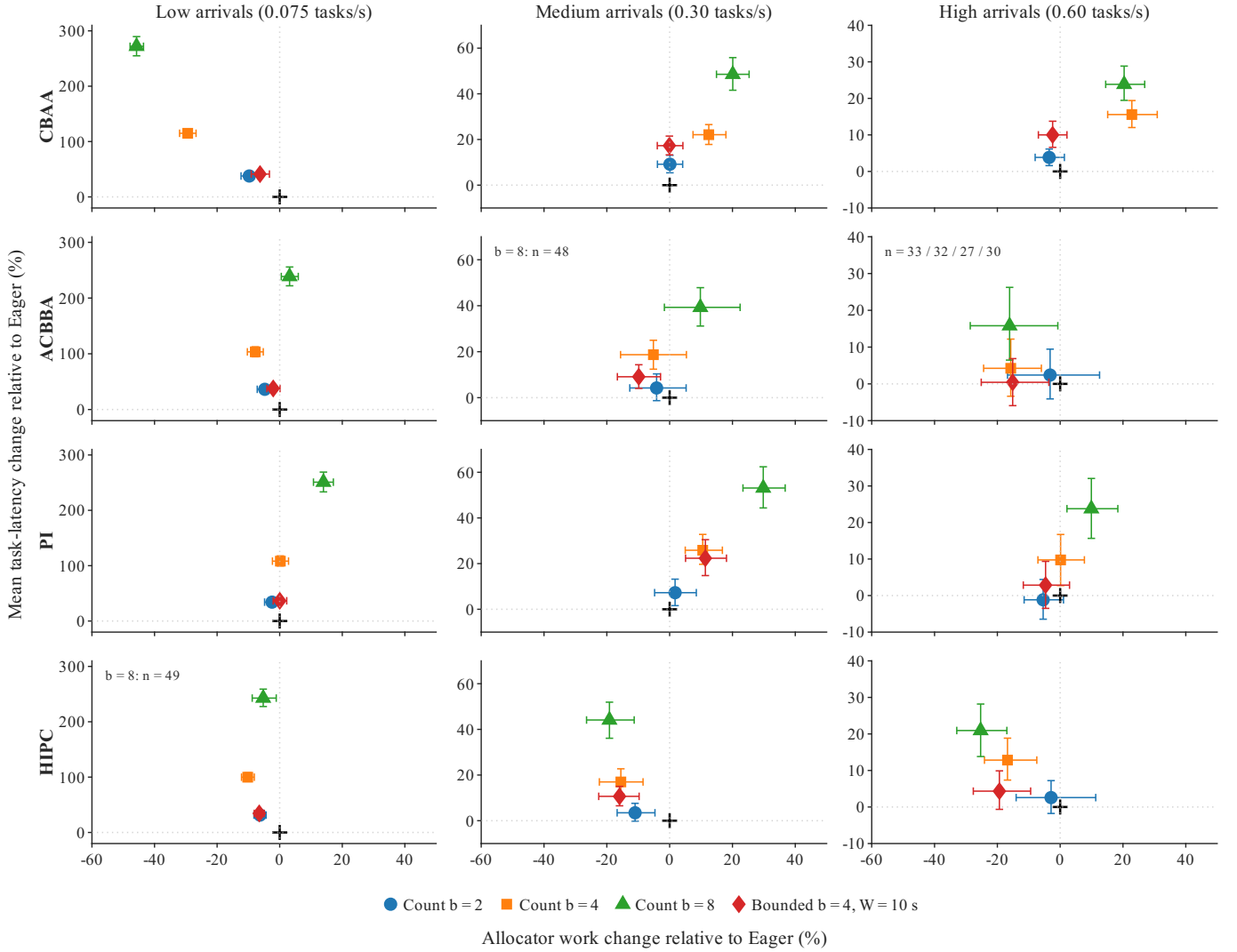}
\caption{Coalescing tradeoff on AGX relative to Eager. The horizontal axis shows allocator-work change and the vertical axis shows mean target service-latency change. Columns indicate arrival loads and rows indicate the representative allocators. Points are percentage changes between paired-cohort means with 95\% pointwise scenario-bootstrap intervals. Negative values indicate lower work or latency. The origin denotes Eager. All comparisons retain 50 pairs except those annotated. The four ACBBA/high sample sizes follow the legend order. Latency scales differ between columns.}
\label{fig:regimes}
\end{figure*}

\subsection{Coalescing Tradeoff Across Arrival Loads}
\label{sec:regimes}
Figure~\ref{fig:regimes} characterizes coalescing through the joint change in allocator processor work and mean target service latency relative to immediate admission. Within each panel, moving left represents computational savings and moving upward represents additional service delay. The upper-left region therefore describes an exchange of responsiveness for lower work, while the upper-right region identifies policies that increase both costs. Comparing the policy trajectories across arrival loads shows how batch size and the waiting deadline change this balance under the representative allocation mechanisms.

Thirty of the 48 comparisons reduce both allocator calls and processor work while increasing mean service latency. Eleven instead increase calls and work despite fewer admission epochs. These are directions of observed mean effects, with uncertainty shown for each contrast. Forty-two comparisons retain all 50 paired scenarios, and the six conditional cohorts are identified in the figure. The predominant computational benefit is thus a tradeoff with target waiting rather than an automatic consequence of reducing admission frequency.

Increasing the Count threshold generally moves the operating point toward greater service delay, but not consistently toward greater work savings. Under sparse arrivals, the eight-task threshold produces the largest latency penalty in every row, while its work response ranges from a substantial reduction to an increase. At higher arrival rates, collecting a batch takes less time and the relative latency penalties become smaller. The computational direction can nevertheless reverse as the workload changes. For the same Count $b=4$ policy, CBAA work falls by 29.39\% under low arrivals but rises by 22.82\% under high arrivals. Threshold selection therefore depends on both the waiting introduced by collection and the work generated after admission.

Sparse arrivals make the waiting component particularly clear. At 0.075 tasks/s, Count $b=4$ increases online-target mean service latency by 20.08--23.57~s across the suite, with an increase on every one of the 50 scenarios for each method. Of this increase, 83.3--96.7\% is attributable to release-to-admission waiting. The observed mean admission wait over the first 40 online targets is 19.51~s, closely matching the 20~s expected from the arrival process and batching policy. The remaining increase occurs after admission during allocation and execution.

Bounded admission limits this waiting component while retaining the opportunity to collect a full batch. For low-load HIPC, online-target mean latency is 14.18~s with Eager, 34.26~s with Count $b=4$, and 21.36~s with Bounded admission. Mean admission wait falls from 19.41 to 7.58~s when the deadline is included. This shifts the operating point toward lower latency without requiring every target to enter allocation immediately. The importance of choosing a waiting window is consistent with online warehouse batching studies~\cite{gilborras2024window}, while Fig.~\ref{fig:regimes} additionally shows the computational response generated after admission.

Across the matrix, Bounded admission lowers mean work in ten of twelve allocator--load conditions with a modest mean service latency increase in all twelve. The work-reducing ACBBA/high comparison contains 30 common completers, while the other nine contain 50. The deadline controls waiting at low arrival load rather than guaranteeing computational savings. Under medium arrivals with PI, for example, Bounded admission increases work by 11.37\% and latency by 22.35\%. Its value is clearest where it retains useful savings while limiting the delay associated with a larger Count threshold. For high-load HIPC, Count $b=8$ reduces work by 25.32\% with a 20.93\% latency increase, whereas Bounded admission reduces work by 19.35\% with a 4.34\% latency increase. The latter retains much of the work reduction at a substantially smaller mean service cost.

The resulting operating regions show that useful coalescing depends jointly on admission waiting and the allocator workload generated after admission. The following analyses examine the source of these workload changes and how the tradeoff shifts under embedded execution.

\subsection{Sources of Computational Savings}
\label{sec:mechanism_results}
The different regions in Fig.~\ref{fig:regimes} arise because coalescing changes both the number and cost of allocator calls. Table~\ref{tab:mechanism} traces two complete high-arrival comparisons from admission to processor work and service latency. Both reduce admission epochs by approximately 72\%, but one reduces the workload generated during execution while the other increases it.

\begin{table}[t]
\caption{High-arrival examples of how coalescing changes allocator work and target-service latency.}
\label{tab:mechanism}
\centering
\footnotesize
\setlength{\tabcolsep}{3.6pt}
\renewcommand{\arraystretch}{1.08}
\begin{tabular}{@{}lrr@{}}
\toprule
Quantity & \shortstack{HIPC\\Bounded $b=4$} & \shortstack{CBAA\\Count $b=4$}\\
\midrule
Admission epochs & $43.00\to12.12$ & $43.00\to12.00$\\
Allocator calls & $1009.90\to835.50$ & $461.08\to495.36$\\
Mean work/call (ms) & $1.631\to1.565$ & $0.304\to0.348$\\
Total allocator work (s) & $1.678\to1.353$ & $0.1405\to0.1726$\\
Mean target-service latency (s) & $32.37\to33.77$ & $27.02\to31.22$\\
\midrule
Call-count contribution (ms) & $-274.29$ & $+11.46$\\
Per-call-work contribution (ms) & $-50.32$ & $+20.61$\\
\bottomrule
\end{tabular}\par
\vspace{3pt}
\noindent\parbox{\columnwidth}{\footnotesize Entries are means over 50 complete paired scenarios and are shown as Eager $\to$ coalescing policy. Bounded uses $W=10$~s. The final two rows decompose the change in total allocator work into contributions associated with call count and mean work per call; they are accounting terms, not separate causal effects.}
\end{table}

With Bounded admission, HIPC allocator work decreases by 19.35\%, with reductions on 38 of 50 paired traces. Allocator calls fall by 17.27\%, and the decomposition in~\eqref{eq:decomposition} attributes 84.5\% of the mean work saving to the call-count term. Mean per-call work changes only slightly, and the work-saving direction is consistent across both 25-scenario execution rounds. Mean target-service latency increases by 1.40~s, yielding a repeatable computational saving with a comparatively modest responsiveness cost.

For Count $b=4$ with CBAA, calls increase by 7.43\% and mean per-call work by 14.20\%, producing a 22.82\% increase in total work, with increases on 42 of 50 paired traces. Both more frequent allocator calls and greater per-call cost contribute to the increase, with the per-call term accounting for the larger share. Mean target-service latency rises by 4.21~s and team travel also increases, indicating that the added processor work accompanies a broader change in allocation and execution rather than directly explaining the full responsiveness penalty. Together with the HIPC result, this shows that the computational effect of coalescing depends on the allocator activity generated after admission, not simply on how many admission events are removed.

\subsection{Embedded Execution}
\label{sec:hardware_results}

The RP2040 experiments test whether the computation--service-latency tradeoff shifts under tighter processor constraints, such that work savings from coalescing become large enough to offset the waiting introduced by batching and improve target service. Count $b=4$ is evaluated against Eager across all four allocators and three arrival loads. The hardware campaign retains 38 complete Eager--Count pairs from 48 planned blocks. Across this matrix, the response is not uniform. Low-load batching increases target-service latency for every allocator, while joint reductions in processor work and service latency emerge in selected medium- and high-load conditions. HIPC at medium load provides the strongest complete example of this shift.

\begin{figure}[t]
\centering
\includegraphics[width=\columnwidth]{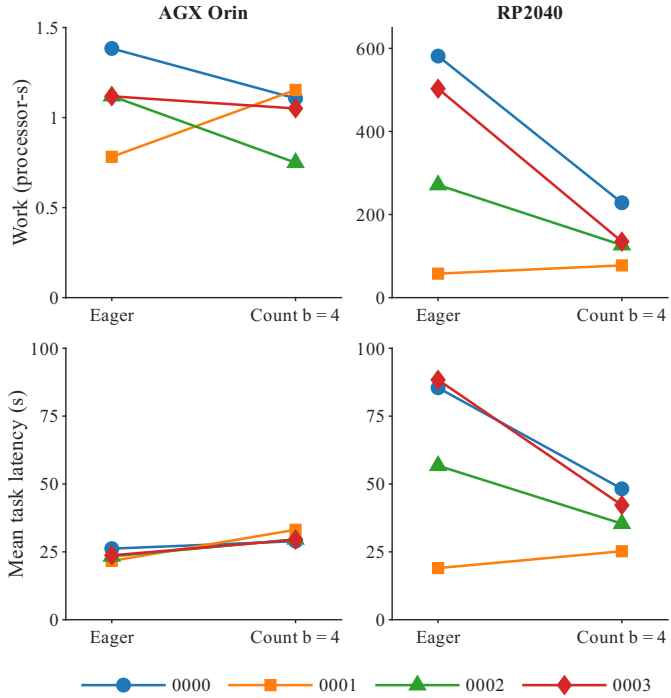}
\caption{Matched medium-arrival HIPC scenarios under Eager and Count $b=4$. Lines connect the same scenario under both policies. The upper panels show summed allocator work on platform-specific scales, and the lower panels show mean target-service latency on a common scale. Legend entries identify the four shared traces.}
\label{fig:hardware}
\end{figure}

Figure~\ref{fig:hardware} shows the four medium-load HIPC scenarios that complete under both policies on both platforms. On RP2040, Count $b=4$ reduces mean allocator work from 353.31 to 141.85 processor-seconds, a 59.85\% reduction, while mean target-service latency falls from 62.39 to 37.76~s, a 39.47\% reduction. Three of the four scenarios improve on both outcomes. The reduction in work reflects both fewer allocator calls and substantially lower work per call. 

The same four AGX scenarios show a different operating point. Mean allocator work decreases by only 7.79\%, while mean target-service latency increases by 27.69\%, with latency increasing on all four traces. Under this condition, coalescing therefore exchanges responsiveness for a modest computational saving on AGX, whereas the larger work reduction on RP2040 is sufficient to offset the waiting introduced by batching and reduce eventual target-service latency. These comparisons use the same admission policy and scenario identities under the execution settings described in Sections~\ref{sec:workloads} and~\ref{sec:platforms}.

The broader RP2040 matrix shows that this shift varies by allocator and target load. PI under high arrivals provides a second favorable case, reducing mean work by 18.67\% and service latency by 12.34\% across its three complete pairs. ACBBA also shows joint work and latency reductions at medium load in its two complete pairs, although the smaller retained cohort due to execution failure could present bias to this result. CBAA does not enter the same favorable region at medium or high load, and sparse-load batching increases service latency for every allocator regardless of whether processor work rises or falls. HIPC itself also changes response at high load, where Count $b=4$ increases mean work by 7.39\%. The embedded results therefore extend the tradeoff in Fig.~\ref{fig:regimes}: tighter processor constraints can make coalescing favorable enough to improve target service, but the crossover depends on both arrival load and the computational response of the allocator.

\subsection{Generality and Deployment Implications}
\label{sec:generality_results}

The zero-compute arm tests whether the workload response to coalescing could be attributed to feedback from processor delay on the AGX Orin. Across the 48 allocator–load–policy contrasts, the direction of the call-count change is preserved in 43 when computation consumes no simulated time. This includes all 12 CBAA and HIPC contrasts, 11 of 12 PI contrasts, and 8 of 12 ACBBA contrasts. Coalescing reduces calls in 36 of 48 causal contrasts and 31 of 48 zero-compute contrasts. The five reversals occur for ACBBA at medium load (Count b = 2 and Bounded) and high load (Count b = 4 and b = 8), and for PI at high load under Bounded admission. In each case, the causal mean decreases while the zero-compute mean increases; the causal changes have magnitudes of only 0.37–4.22\%, consistent with computation-delay feedback shifting conditions near the boundary between increased and reduced allocator activity.

The latency results likewise reflect the admission mechanism rather than the terminal residual alone. Restricting the analysis to the first 40 online targets preserves the principal latency effects, including the sparse-arrival penalties, thereby removing the final residual from the comparison. Together with the embedded results, these checks show when coalescing can be useful. Batching can reduce the computational disturbance created by dynamic task arrivals, and that reduction becomes increasingly consequential as allocator execution occupies a larger fraction of the mission timescale. Its benefit is therefore greatest when batching meaningfully reduces downstream allocator activity without introducing excessive admission waiting.

The measured operating regions apply to four robots, exponential arrivals, lossless communication, and the evaluated implementations. Autonomous trigger rules, state retention, arrival structure, and processor capability could shift the quantitative boundary between useful computational savings and added target waiting. Hardware results measure allocator execution within simulated missions rather than physical robot motion. Future work should test whether the observed operating regions persist with larger teams, degraded communication, non-Poisson arrivals, and fully embodied hardware. More broadly, these extensions support the goal of making decentralized MRTA more computationally efficient and deployable on resource-constrained edge platforms.

\section{Conclusion}
\label{sec:conclusion}

This study evaluates task-admission coalescing as a mechanism for reducing the computational demand created by dynamically released tasks in decentralized multi-robot teams. Across four MRTA allocators, coalescing lowers allocator work in most evaluated conditions, showing that batching can meaningfully reduce the computational burden of online task arrivals. The magnitude of that saving depends on how each allocator responds after admission, rather than on admission count alone. This behavior is largely preserved when computation is removed from simulated time, with the direction of the call-count response unchanged in 43 of 48 contrasts. The main cost appears under sparse arrivals, where waiting for a batch to form can dominate the added target-service latency, while bounded admission limits that delay. Under high-arrival HIPC on AGX, Bounded admission reduces mean allocator work by 19.35\% with an observed 1.40-s increase in mean target-service latency.

The embedded experiments show why this mechanism can become more useful on resource-constrained, embedded devices. Across four medium-arrival HIPC scenarios on RP2040, Count $b=4$ reduces mean allocator work by 59.85\% and target-service latency by 39.47\%, with joint improvements also appearing for other allocator--load conditions. However, this result is not universal among allocators and target loads, demonstrating that usefulness depends on whether batching actually reduces the downstream computational workload and whether that saving is large relative to the waiting it introduces. Task admission therefore provides a practical system-level mechanism for moderating the computational impact of dynamic task arrivals, particularly on resource-constrained platforms where allocator execution time can materially affect target service.

\bibliographystyle{IEEEtran}
\bibliography{reference}

@misc{lott2026benchmark,
  author = {Lott, James and Honary, Vahraz},
  title = {Decentralized Multi-Robot Task Allocation Under Degraded Communication: A Benchmark of Performance, Reliability, and Computation},
  year = {2026},
  eprint = {2609.13711},
  archivePrefix = {arXiv},
  note = {arXiv:2609.13711},
  url = {https://arxiv.org/abs/2609.13711}
}

@misc{lott2026bundle,
  author = {Lott, James and Honary, Vahraz},
  title = {Bundle Length Tradeoffs in Decentralized Multi-Robot Task Allocation Under Degraded Communications},
  year = {2026},
  eprint = {2609.06273},
  archivePrefix = {arXiv},
  note = {arXiv:2609.06273},
  url = {https://arxiv.org/abs/2609.06273}
}

@article{cao2025cv,
  author = {Cao, Yan and Long, Teng and Sun, Jingliang and Wang, Zhu and Xu, Guangtong},
  title = {Comparison of Distributed Task Allocation Algorithms Considering Non-Ideal Communication Factors for Multi-{UAV} Collaborative Visit Missions},
  year = {2025},
  doi = {10.1109/LRA.2023.3295999},
  journal = {IEEE Robotics and Automation Letters},
  volume = {10},
  number = {2},
  pages = {1928--1935}
}

@article{choi2009consensus,
  author = {Choi, Han-Lim and Brunet, Luc and How, Jonathan P.},
  title = {Consensus-Based Decentralized Auctions for Robust Task Allocation},
  year = {2009},
  doi = {10.1109/TRO.2009.2022423},
  journal = {IEEE Transactions on Robotics},
  volume = {25},
  number = {4},
  pages = {912--926}
}

@inproceedings{johnson2010asynchronous,
  author = {Johnson, Luke B. and Ponda, Sameera S. and Choi, Han-Lim and How, Jonathan P.},
  title = {Improving the Efficiency of a Decentralized Tasking Algorithm for {UAV} Teams with Asynchronous Communications},
  year = {2010},
  doi = {10.2514/6.2010-8421},
  booktitle = {AIAA Guidance, Navigation, and Control Conference}
}

@inproceedings{johnson2011dynamic,
  author = {Johnson, Luke B. and Ponda, Sameera S. and Choi, Han-Lim and How, Jonathan P.},
  title = {Asynchronous Decentralized Task Allocation for Dynamic Environments},
  year = {2011},
  doi = {10.2514/6.2011-1441},
  booktitle = {Infotech@Aerospace 2011}
}

@incollection{johnson2016hipc,
  author = {Johnson, Luke and Choi, Han-Lim and How, Jonathan P.},
  title = {The Hybrid Information and Plan Consensus Algorithm with Imperfect Situational Awareness},
  year = {2016},
  doi = {10.1007/978-4-431-55879-8_16},
  booktitle = {Distributed Autonomous Robotic Systems},
  publisher = {Springer},
  series = {Springer Tracts in Advanced Robotics},
  volume = {112},
  pages = {221--233}
}

@article{zhao2016impact,
  author = {Zhao, Wanqing and Meng, Qinggang and Chung, Paul W. H.},
  title = {A Heuristic Distributed Task Allocation Method for Multivehicle Multitask Problems and Its Application to Search and Rescue Scenario},
  year = {2016},
  doi = {10.1109/TCYB.2015.2418052},
  journal = {IEEE Transactions on Cybernetics},
  volume = {46},
  number = {4},
  pages = {902--915}
}

@inproceedings{buckman2019partial,
  author = {Buckman, Noam and Choi, Han-Lim and How, Jonathan P.},
  title = {Partial Replanning for Decentralized Dynamic Task Allocation},
  year = {2019},
  doi = {10.2514/6.2019-0915},
  booktitle = {AIAA Scitech 2019 Forum}
}

@article{pavone2011adaptive,
  author = {Pavone, Marco and Frazzoli, Emilio and Bullo, Francesco},
  title = {Adaptive and Distributed Algorithms for Vehicle Routing in a Stochastic and Dynamic Environment},
  year = {2011},
  doi = {10.1109/TAC.2010.2092850},
  journal = {IEEE Transactions on Automatic Control},
  volume = {56},
  number = {6},
  pages = {1259--1274}
}

@article{aznar2026event,
  author = {Aznar, Fidel and Pujol, Mar and D{\'i}ez, {\'A}lvaro},
  title = {Event-Triggered Adaptive Consensus for Multi-Robot Task Allocation},
  year = {2026},
  doi = {10.1016/j.comcom.2026.108499},
  journal = {Computer Communications},
  volume = {251},
  pages = {108499}
}

@inproceedings{sao2025event,
  author = {Sao, Vinita and Ho, Tu Dac and Bhore, Sujoy and Sujit, P. B.},
  title = {Event Driven {CBBA} with Reduced Communication},
  booktitle = {2025 International Conference on Unmanned Aircraft Systems (ICUAS)},
  year = {2025},
  pages = {504--510},
  doi = {10.1109/ICUAS65942.2025.11007849}
}

@inproceedings{tuck2024smt,
  author = {Tuck, Victoria Marie and Chen, Pei-Wei and Fainekos, Georgios and Hoxha, Bardh and Okamoto, Hideki and Sastry, S. Shankar and Seshia, Sanjit A.},
  title = {{SMT}-Based Dynamic Multi-Robot Task Allocation},
  booktitle = {{NASA} Formal Methods},
  series = {Lecture Notes in Computer Science},
  volume = {14627},
  pages = {331--351},
  publisher = {Springer},
  year = {2024},
  url = {https://arxiv.org/abs/2403.11737}
}

@article{efron1979bootstrap,
  author = {Efron, Bradley},
  title = {Bootstrap Methods: Another Look at the Jackknife},
  journal = {The Annals of Statistics},
  volume = {7},
  number = {1},
  pages = {1--26},
  year = {1979},
  doi = {10.1214/aos/1176344552}
}

@article{gilborras2024window,
  author = {Gil-Borr{\'a}s, Sergio and Pardo, Eduardo G. and Jim{\'e}nez, Ernesto and S{\"o}rensen, Kenneth},
  title = {The Time-Window Strategy in the Online Order Batching Problem},
  journal = {International Journal of Production Research},
  volume = {62},
  number = {12},
  pages = {4446--4469},
  year = {2024},
  doi = {10.1080/00207543.2023.2263884}
}

@INPROCEEDINGS{mypaper1,
  author={Lott, James and Honary, Vahraz},
  booktitle={2026 IEEE 16th Annual Computing and Communication Workshop and Conference (CCWC)}, 
  title={Lightweight Bidding Framework for Communication-Efficient Multi-Robot Coordination in Clue-Based Search}, 
  year={2026},
  volume={},
  number={},
  pages={1214-1220},
  doi={10.1109/CCWC67433.2026.11393706}}

\end{document}